\documentclass{article} 
\usepackage{nips13submit_e,times}
\usepackage{hyperref}
\usepackage{url}

\usepackage{graphicx}
\usepackage{amsmath}
\usepackage{amssymb}
\usepackage{caption}
\usepackage{subcaption}

\title{Learning to Generate Networks}

\author{
James Atwood \\
School of Computer Science\\
University of Massachusetts Amherst\\
Amherst, MA\\
\texttt{jatwood@cs.umass.edu} \\
\And
Don Towsley \\
School of Computer Science\\
University of Massachusetts Amherst\\
Amherst, MA\\
\texttt{towsley@cs.umass.edu} \\
\And
Krista Gile \\
Department of Mathematics and Statistics \\
University of Massachusetts Amherst\\
Amherst, MA\\
\texttt{gile@math.umass.edu} \\
\And
David Jensen \\
School of Computer Science\\
University of Massachusetts Amherst\\
Amherst, MA\\
\texttt{jensen@cs.umass.edu} \\
}

\nipsfinalcopy

\newcommand{\netset}{\mathcal{N}}

\begin{document}

\maketitle

\begin{abstract}
  We investigate the problem of learning to generate complex networks
  from data.  Specifically, we consider whether deep belief networks, dependency
  networks, and members of the exponential random graph family can
  learn to generate networks whose complex behavior is consistent with
  a set of input examples.  We find that the deep model is able to
  capture the complex behavior of small networks, but that no model is
  able capture this behavior for networks with more than a handful of
  nodes.


\end{abstract}

\section{Introduction}

Researchers in network science have long studied the behavior of
synthetic models of network structure whose intent is to explain how
an observed network behavior arises. For instance, Price's model
demonstrates that the heavy-tailed degree distribution observed in
citation networks can be thought of as a consequence of a preferential
attachment process \cite{Price:1976vn}.  Generally speaking, a
researcher defines a generative process (such as preferential
attachment), generates multiple networks via that process, and then
determines whether the generated networks ``look like'' real networks.

This work is concerned with reversing the process described above.
Can a generative process be learned from data?  Can a single model
encode many different generative processes?  That is, rather than
accounting for a specific complex behavior through a predefined
process, the model should be able to represent as broad a set of
networks as possible with minimal input from a user.

The goal of this work is to find a flexible model that is able to
learn a generative network process from data.  We consider whether
models such as deep belief networks, dependency networks, and members
of the exponential random graph family can learn to generate networks
whose complex behavior is consistent with a set of input examples.  We
find that deep belief networks are the only model to perform well at
any network size, and we find that all models perform poorly when
generating moderate to large networks.  The good performance of the
deep model for small network sizes suggests that it is learning latent
representations that encode limited complex network behavior.

The remainder of the paper is structured as follows.  Section
\ref{sec:problem} presents the task and a set of candidate models.
Section \ref{sec:assessment} contains an evaluation method for the
experiments described in Section \ref{sec:experiments}.  Results and
conclusions are discussed in Sections \ref{sec:results} and
\ref{sec:conclusion}, respectively.

%

\section{Problem Formulation}
\label{sec:problem}
Consider a fixed set of nodes $V$ and an adjacency matrix
$\mathbf{Y}$, where each element $Y_{ij}$ takes on value 1 if there is
an edge from node $i$ to node $j$ and 0 otherwise.  Given a set of
observed network examples, the task is to learn a joint distribution
over the $|V|^2$ binary random variables that comprise $\mathbf{Y}$ in
such a way that complex network structure is maintained.  For
instance, if the input consists of scale-free networks that were
generated from a preferential attachment process, the output networks
should be scale-free as well.  Note that this is a substantial
departure from most work in network generation, which focuses on
either understanding a predefined process that is not learned from
data, or estimating potential generative processes from a single
network instance rather than many.

A naive method would be to place no constraints on the dependencies
between edges; that is, to model $\text{P}(\mathbf{Y} = \mathbf{y})$
as a fully-connected Markov random field (MRF).  Learning and exact
inference in this model are computationally intractable, however.  We
could mitigate this intractability by learning the dependence
relationships from data; that is, learning the structure of the MRF.
However, learning the structure of undirected discrete graphical
models is itself intractable \cite[Chapter~26.8]{mpp:38136}.

An alternative approach is to learn \emph{approximate} models; that
is, models that provide tractable inference and learning of an
approximation to the joint distribution.  We present an application of
two such models: dependency networks and deep belief networks.

\subsection{Dependency Networks}
A dependency network \cite{Heckerman:2000dc} is a form of
probabilistic graphical model.  A dependency network can be
thought of as a collection of regressions or classifications among
variables that can be combined via Gibbs sampling to define a joint
distribution.  This representation facilitates straightforward and
efficient algorithms for structure learning and inference.

A dependency network is constructed over a set of variables $D$ (in
this case, the elements of the adjacency matrix) by independently fitting a sparse
full-conditional distribution to each variable
\cite[Chapter~26.2.2]{mpp:38136}.  Any available method for fitting
these sparse conditional distributions can be selected; Heckerman et
al. use probabilistic decision trees and support vector machines, for
instance \cite{Heckerman:2000dc}.  The graphical component of the
dependency network is a bidirected unweighted network in which the
elements of $D$ are nodes, and there is an edge from $d_1 \in D$ to
$d_2 \in D$ if $d_2$'s conditional model depends on $d_1$.

We choose L1-penalized logistic regression for learning the sparse
conditional models due to its appropriateness for binary data,
conceptual and computational simplicity, and readily available
implementations \cite{scikit-learn}.  With this choice
of model, the dependency network can be represented as a
collection of conditional distributions
\begin{equation}
  P(Y_{ij} = 1 | \mathbf{Y} \setminus \left\{Y_{ij}\right\}) =
  \frac{\text{exp}\left(\beta_k \cdot \mathbf{Y} \setminus \left\{Y_{ij}\right\}\right)}
       {1 + \text{exp}\left(\beta_k \cdot \mathbf{Y} \setminus \left\{Y_{ij}\right\}\right)}
\end{equation}

\subsection{Restricted Boltzmann Machines}
A Boltzmann machine is a variety of Markov Random Field with a set of
hidden (latent) nodes $\mathbf{h}$ and a set of visible (observable)
nodes $\mathbf{x}$ \cite{ackley1985learning},
\cite[Chapter~27.7]{mpp:38136}.  In this work, the visible variables
are the elements of the adjacency matrix.  Exact inference in a
Boltzmann machine is intractable and approximate inference can be slow
\cite[Chapter~27.7]{mpp:38136}.  The speed of inference can be
improved by only allowing edges between elements of $\mathbf{h}$ and
$\mathbf{x}$; such a model is known as a restricted Boltzmann machine
(RBM).  We consider binary RBMs which constrain the hidden and visible
variables to be binary-valued \cite{hinton2002training}.  One can
interpret the hidden variables in a binary RBM as `gates' which govern
first-order correlations between visible variables.

More formally, a binary restricted Boltzmann machine is defined as an
undirected fully-connected bipartite graph with `visible' nodes
$\pmb{x}$ and `hidden' nodes $\pmb{h}$, and a $|\pmb{x}| \times
|\pmb{h}|$ real matrix of parameters $W$.  This model defines a joint
distribution
\begin{equation}
  P(\pmb{x},\pmb{h}) = \mathcal{Z}^{-1} \text{exp}\left(\pmb{x}^T W \pmb{h}\right)  
\end{equation}
where $\mathcal{Z} = \sum_{\pmb{x'}, \pmb{h'}}
\text{exp}\left(\pmb{x'}^T W \pmb{h'}\right)$ is the normalization
constant.  The probability of the visible variables $\pmb{x}$ is found
by marginalizing over the hidden variables $\pmb{h}$
\begin{equation}
  P(\pmb{x}) = \mathcal{Z}^{-1} \sum_{\pmb{h'}} \text{exp}\left(\pmb{x}^T W \pmb{h'}\right)
\end{equation}
There are many available methods for learning the parameters ($W$) of
restricted Boltzmann machines.  We select the stochastic maximum
likelihood method described by Marlin et al. due to its `consistently
better performance in terms of density estimation'
\cite{Marlin:2010tx}.

\subsection{Deep Belief Networks}
Deep belief networks (DBNs) can be thought of as stacked RBMs.  We can
think of the higher levels of the deep belief network as gates which
mitigate higher-order correlations between visible variables.  This
allows deep belief networks to represent very complex dependence
structures.  However, exact learning and inference for such a
model is highly intractable.

Fortunately, Hinton, Osindero, and Teh introduced a greedy approximate
algorithm for tractably learning the parameters of this
structure \cite{Hinton:2006tm}.  Each layer of the hierarchical
structure is learned greedily; that is, as its own restricted
Boltzmann machine.  Data at higher levels are inferred using the
parameters learned at lower levels.  This approximate method has been
very successful in practice, providing state-of-the-art performance
for the classification of handwritten digits, for instance
\cite{Hinton:2006tm}.  The algorithm is theoretically justified using
the idea of complementary priors; for more details, please see `A fast
learning algorithm for deep belief nets' Appendix A
\cite{Hinton:2006tm}.

\subsection{Exponential Random Graph Models}

An exponential random graph model (ERGM) models a network as a
collection of edges whose probabilities are given as a normalized
exponential function of configurations.  Configurations can be
constructed from any collection of dyads; some examples include a
single edge, a reciprocated pair of edges, or three edges forming a
cycle.  Configurations must be specified by the experimenters.  The
form of the model is
\begin{equation}
  \text{P}(\mathbf{Y} = \mathbf{y}) = \left(\frac{1}{\kappa}\right) \text{exp}\left\{\sum\limits_{A}\eta_Ag_A(\mathbf{y})\right\} \\
\end{equation}
where the summation is over all configurations, $\eta_A$ is the
parameter corresponding to configuration $A$, $g_A(\mathbf{y}) =
\prod_{y_{ij} \in A} y_{ij}$ is the network statistic corresponding to
$A$, and $\kappa$ is the normalization constant.  The parameter
$\eta_A$ can be interpreted as an index of the level of configuration
$A$.  For a more complete treatment, please see `An introduction to
exponential random graph ($p^*$) models for social networks' by Robins
et al \cite{Robins:2007kb}.


\begin{table}
  \centering
  \begin{tabular}{|l|c|c|c|}
    \hline
    configuration & p1 & markov & higher-order \\
    \hline
    \# edges                    & X & X & X \\
    \# reciprocated edges       & X & X & X \\
    two in stars                & & X & X \\
    two out stars               & & X & X \\
    two mixed stars             & & X & X \\
    transitive triads           & & X & X \\
    geometric sum of in-degree  & & & X \\
    geometric sum of out-degree & & & X \\
    two-paths                   & & & X \\
    alternating k-triangles     & & & X \\
    alternating k-paths         & & & X \\
    \hline
  \end{tabular}
  \caption{ERGM model specification.}
  \label{tab:ergm}
\end{table}
The ERGM framework accommodates a family of models of varying
complexity.  The simplest model assumes that all edges are
independent.  This is known as the Bernoulli graph model
\cite{Robins:2007kb} or the Erdos-Renyi (ER) model
\cite{ERDdS:1959ud}.  A slightly more complex formulation is found in
the $p_1$ model, which assumes that \emph{dyads} are independent of
each other.  The $p_2$ model assumes a looser dyadic independence,
where dyads are independent given node-specific attributes
\cite{Lazega:1997us,Robins:2007kb}.


Markov random graphs extend these dependence notions.  In the model of
Frank and Strauss, an edge $Y_{ij}$ depends on any other edge
involving $i$ or $j$, conditioned on the rest of the graph \cite{Frank:1986va}.  This
implies a set of possible configurations involving two or three nodes;
please see Figure 1 of \cite{Robins:2007kb} for more details.

Finally, there is the higher-order model of Snijders and collaborators
\cite{snijders2006new, Robins:2007el}, which includes a higher order
of dependencies than those that can be represented by Markov random
graphs.  In contrast to other methods, the higher-order specification
includes statistics derived from the entire network, such as a
geometrically-weighted sum of the distribution of in-degree.

In this work, we consider the $p_1$, Markov random, and new
specification graph models.  We omit $p_2$ because this work is not
concerned with nodal attributes.  Table \ref{tab:ergm} specifies the
configurations used in each model.  It should be noted that we did not
include the ``fixed effects'' component of the $p_1$ model.

It should also be noted that members of the exponential random graph
model are not typically used to generate networks.  In fact, some
ERGMs have been observed to exhibit `near degeneracy'.  A graph
distribution $\text{P}(\mathbf{Y} = \mathbf{y})$ is termed
`near-degenerate' if it puts most of its probability mass on only a
few (possibly one or two) distinct graphs
\cite{handcock2003statistical, handcock2003assessing,
  Schweinberger:2011fb}.  The implied graphs are usually highly
unrealistic; for instance, fully-connected meshes or graphs with no
edges at all.  This property has been observed across members of the
ERGM family, from the relatively uncomplicated Markov graph model of
Frank and Strauss \cite{Robins:2007el} to the higher order model of
Snijders and collaborators \cite{Schweinberger:2011fb,
  snijders2006new}, although the higher-order models do a better job
of avoiding it.

ERGMs are also typically fit to a single network rather than a set.
We adapt ERGMs to the multiple input setting by assuming that the input
networks are independent and identically distributed and maximizing
their joint likelihood.

\begin{table*}
  \centering
  
  \tiny\begin{tabular}{|c||c|c||c|c||c|c||c|c||c|c|}
    \hline
    \multicolumn{11}{c}{\textbf{Directed ER}} \\
    \hline
    & \multicolumn{2}{|c||}{deep} & \multicolumn{2}{|c||}{dependency} & \multicolumn{2}{|c||}{p1} & \multicolumn{2}{|c||}{markov} & \multicolumn{2}{|c|}{higher-order}\\
    \hline
    Network Size & $\bar{\mu}_d$ & $\bar{\mu}_c$ & $\bar{\mu}_d$ & $\bar{\mu}_c$ & $\bar{\mu}_d$ & $\bar{\mu}_c$ & $\bar{\mu}_d$ & $\bar{\mu}_c$ & $\bar{\mu}_d$ & $\bar{\mu}_c$\\    
    \hline
    4 & \textbf{0.2810} & \textbf{0.0520} & 0.0000 & 0.0000 & 0.0000 & \textbf{0.3290} & \textbf{0.4690} & 0.0260 & \textbf{0.4430} & 0.0020  \\
    6 & 0.0140 & \textbf{0.0550} & 0.0000 & 0.0000 & \textbf{0.5880} & 0.0000 & 0.0020 & 0.0010 & 0.0040 & \textbf{0.5920}  \\
    8 & \textbf{0.6390} & \textbf{0.9350} & 0.0020 & 0.0000 & 0.0000 & 0.0000 & 0.0000 & 0.0000 & 0.0000 & 0.0000  \\
    10 & \textbf{0.5310} & \textbf{0.7860} & 0.0000 & \textbf{0.3330} & 0.0000 & 0.0000 & 0.0000 & 0.0000 & 0.0000 & 0.0000  \\
    20 & 0.0000 & 0.0000 & \textbf{0.0650} & \textbf{0.7640} & 0.0000 & 0.0000 & 0.0000 & 0.0000 & 0.0000 & 0.0000  \\
    \hline
  \end{tabular}

  \begin{tabular}{|c||c|c||c|c||c|c||c|c||c|c|}
    \hline
    \multicolumn{11}{c}{\textbf{Configuration Model}} \\
    \hline
    & \multicolumn{2}{|c||}{deep} & \multicolumn{2}{|c||}{dependency} & \multicolumn{2}{|c||}{p1} & \multicolumn{2}{|c||}{markov} & \multicolumn{2}{|c|}{higher-order}\\
    \hline
    Network Size & $\bar{\mu}_d$ & $\bar{\mu}_c$ & $\bar{\mu}_d$ & $\bar{\mu}_c$ & $\bar{\mu}_d$ & $\bar{\mu}_c$ & $\bar{\mu}_d$ & $\bar{\mu}_c$ & $\bar{\mu}_d$ & $\bar{\mu}_c$\\    
    \hline
    4 & \textbf{0.0630} & 0.0400 & \textbf{1.0000} & 0.0000 & 0.0000 & 0.0000 & 0.0000 & \textbf{0.5260} & 0.0010 & \textbf{0.8510}  \\
    6 & 0.0000 & 0.0000 & 0.0000 & 0.0000 & 0.0000 & 0.0020 & \textbf{0.4170} & 0.0350 & \textbf{0.4190} & 0.0410  \\
    8 & 0.0000 & 0.0130 & 0.0000 & 0.0000 & 0.0000 & 0.0000 & 0.0000 & 0.0000 & 0.0000 & 0.0000  \\
    10 & 0.0000 & \textbf{0.2250} & 0.0000 & 0.0000 & 0.0000 & 0.0000 & 0.0000 & 0.0000 & 0.0000 & 0.0000  \\
    20 & 0.0000 & 0.0000 & 0.0000 & 0.0010 & 0.0000 & 0.0000 & 0.0000 & 0.0000 & 0.0000 & 0.0000  \\
    \hline
  \end{tabular}

  \begin{tabular}{|c||c|c||c|c||c|c||c|c||c|c|}
    \hline
    \multicolumn{11}{c}{\textbf{Krapivsky's Model}} \\
    \hline
    & \multicolumn{2}{|c||}{deep} & \multicolumn{2}{|c||}{dependency} & \multicolumn{2}{|c||}{p1} & \multicolumn{2}{|c||}{markov} & \multicolumn{2}{|c|}{higher-order}\\
    \hline
    Network Size & $\bar{\mu}_d$ & $\bar{\mu}_c$ & $\bar{\mu}_d$ & $\bar{\mu}_c$ & $\bar{\mu}_d$ & $\bar{\mu}_c$ & $\bar{\mu}_d$ & $\bar{\mu}_c$ & $\bar{\mu}_d$ & $\bar{\mu}_c$\\    
    \hline
    4 & \textbf{0.5480} & 0.0000 & 0.0000 & \textbf{1.0000} & 0.0000 & 0.0000 & \textbf{0.1760} & 0.0000 & \textbf{0.1710} & 0.0000  \\
    6 & \textbf{0.1210} & 0.0120 & 0.0000 & 0.0060 & 0.0000 & 0.0300 & 0.0000 & 0.0000 & 0.0000 & 0.0000  \\
    8 & \textbf{0.7760} & \textbf{0.2760} & \textbf{0.3990} & \textbf{0.0670} & \textbf{0.7450} & 0.0000 & 0.0000 & 0.0300 & 0.0000 & \textbf{0.4590}  \\
    10 & \textbf{0.3870} & 0.0150 & 0.0000 & 0.0000 & 0.0070 & \textbf{0.5290} & \textbf{0.1830} & \textbf{0.8160} & \textbf{0.1770} & \textbf{0.3800}  \\
    20 & 0.0010 & \textbf{0.4290} & 0.0000 & 0.0000 & 0.0000 & 0.0000 & 0.0000 & 0.0000 & 0.0000 & 0.0000  \\
    \hline
  \end{tabular}
  \begin{tabular}{|c||c|c||c|c||c|c||c|c||c|c|}
    \hline
    \multicolumn{11}{c}{\textbf{Twitter}} \\
    \hline
    & \multicolumn{2}{|c||}{deep} & \multicolumn{2}{|c||}{dependency} & \multicolumn{2}{|c||}{p1} & \multicolumn{2}{|c||}{markov} & \multicolumn{2}{|c|}{higher-order}\\
    \hline
    Network Size & $\bar{\mu}_d$ & $\bar{\mu}_c$ & $\bar{\mu}_d$ & $\bar{\mu}_c$ & $\bar{\mu}_d$ & $\bar{\mu}_c$ & $\bar{\mu}_d$ & $\bar{\mu}_c$ & $\bar{\mu}_d$ & $\bar{\mu}_c$\\    
    \hline
    4 & 0.0000 & \textbf{0.0770} & 0.0000 & 0.0000 & 0.0000 & 0.0000 & 0.0170 & \textbf{0.2480} & 0.0190 & \textbf{0.1230}  \\
    6 & \textbf{0.6030} & \textbf{0.6650} & \textbf{0.0590} & 0.0000 & 0.0000 & 0.0060 & 0.0000 & \textbf{0.5550} & 0.0000 & 0.0000  \\
    8 & \textbf{0.5950} & \textbf{0.3090} & 0.0000 & 0.0010 & 0.0000 & 0.0020 & 0.0000 & 0.0000 & 0.0000 & 0.0000  \\
    10 & \textbf{0.1220} & \textbf{0.2910} & 0.0000 & 0.0000 & \textbf{0.9160} & \textbf{0.0750} & \textbf{0.1350} & \textbf{0.3200} & \textbf{0.4690} & \textbf{0.2610}  \\
    20 & 0.0000 & 0.0000 & 0.0000 & 0.0020 & 0.0000 & 0.0000 & 0.0000 & 0.0000 & 0.0010 & 0.0000  \\
    \hline
  \end{tabular}
  \caption{\small The application of the permutation test describe in Section
    \ref{sec:perm}.  Each column lists the $p$-value of the
    permutation test for a particular generative model and network
    statistic when applied to networks of different
    sizes. \textbf{Bold} values indicate that we can not reject the
    null hypothesis that the networks are indistinguishable at
    significance $\alpha=0.05$; i.e. that the model succeeded in
    reproducing the ground-truth networks with respect to this
    statistic.  $\bar{\mu}_d$ is the mean in-degree and $\bar{\mu}_c$
    the mean clustering coefficient.  Note that the deep model is most
    consistently able to reproduce the data.}
    \label{tab:synthperm}
\end{table*}

\section{Assessing the Quality of Generated Networks}
\label{sec:assessment}

The task of this work is to generate samples that `look like' data.
Ideally, we could formulate a testable hypothesis that one set of
networks looks like another without appealing to the models that
generated them.  In the section below we describe a method for testing
a minimum condition for effective network generation.



\subsection{Permutation Test}
\label{sec:perm}
For each network, we estimate a mean from a set of values
corresponding to the nodes in the network.  We then test whether the
distribution of those estimated mean values from a set of generated
networks is significantly different from those of the original
networks from which the generator was learned.  If this condition is
satisfied, we still may not be learning good generative processes, but
failing this condition implies that we are learning poor generators.

More formally, let $\Omega$ be the sample space defined by the set of
all networks with $|V|$ nodes and let $f: \Omega \rightarrow
\mathbb{R}^d$ be a function which maps a network to a real vector;
i.e., $f$ defines a vector of random variables.  Let $N$ be some
observed network, let $X_N$ be the result of the application of $f$ to
$N$, and $\mu_{X_N}$ be the mean of $X$.  For instance, if $f$ returns
the degree distribution of a network, then $\mu_{X_N}$ is the mean
degree of $N$.

Consider two sets of networks $\netset_1$ and $\netset_2$; let
$\pmb{\mu}_{\mathcal{X}_i}$ be the mean of the mean values obtained by
applying $f$ to network set $\netset_i$.  For instance, if $f$ returns
the degree distribution of a network, then $\pmb{\mu}_{\mathcal{X}_1}$
is the mean of all of the mean degrees of $\netset_1$.

We wish to assess the hypothesis that
sample means $\pmb{\mu}_{\mathcal{X}_1}$ and
$\pmb{\mu}_{\mathcal{X}_2}$ are drawn from the same distribution.  We
assess this by computing an observed test statistic $t_{obs} =
|\pmb{\mu}_{\mathcal{X}_1} - \pmb{\mu}_{\mathcal{X}_2}|$ and a set of
$n$ `permuted' test statistics $t_{perm} =
\left\{|\pmb{\mu}_{\mathcal{X}^k_1} - \pmb{\mu}_{\mathcal{X}^k_2}| \:
: k = 1,2,...,n\right\}$ where $\pmb{\mu}_{\mathcal{X}^k_1}$ and
$\pmb{\mu}_{\mathcal{X}^k_2}$ are the sample means obtained by
repartitioning $\netset_1$ and $\netset_2$ according to some label
permutation $k$.

Let $p$ be the proportion of permuted means with a larger absolute
difference than the true absolute difference.  If $p\geq\alpha$, we
conclude that we can not reject the null hypothesis that the means are
drawn from the same distribution at some significance level $\alpha$
\cite{dwass1957}, implying that $\netset_1$ and $\netset_2$ are not
distinguishable with respect to $f$.

This test tells us whether the generative model is able to learn the
first moment of a specified complex network behavior (such as the
degree distribution).  As previously mentioned, if this condition is
satisfied, we still may not be learning good generative processes
(because the generated networks may not properly represent the higher
moments of the complex behavior of the training networks), but failure
does imply that we are learning poor generators.

%

\begin{table*}
  \centering
  \small\begin{tabular}{|c|c|c|}
    \hline
    Network Size & $\bar{\mu}_d$ & $\bar{\mu}_c$ \\
    \hline
    4 & \textbf{0.5480} & 0.0000 \\
    6 & \textbf{0.6030} & \textbf{0.6650} \\
    8 & \textbf{0.5950} & \textbf{0.3090} \\
    10 & \textbf{0.1220} & \textbf{0.2910} \\
    20 & 0.0000 & 0.0000 \\
    100 & 0.0000 & 0.0000 \\
    200 & 0.0000 & \textbf{0.3160} \\
    400 & 0.0000 & 0.0000 \\
    600 & 0.0000 & 0.0000  \\
    \hline
  \end{tabular}
  \caption{\small The permutation test applied to larger networks generated
    by the deep belief network.  Note that the performance of the deep
    model degrades sharply as the networks increase in size.}
  \label{tab:kraplarge}
\end{table*}

\begin{figure*}
  \centering
  \includegraphics[scale=0.5]{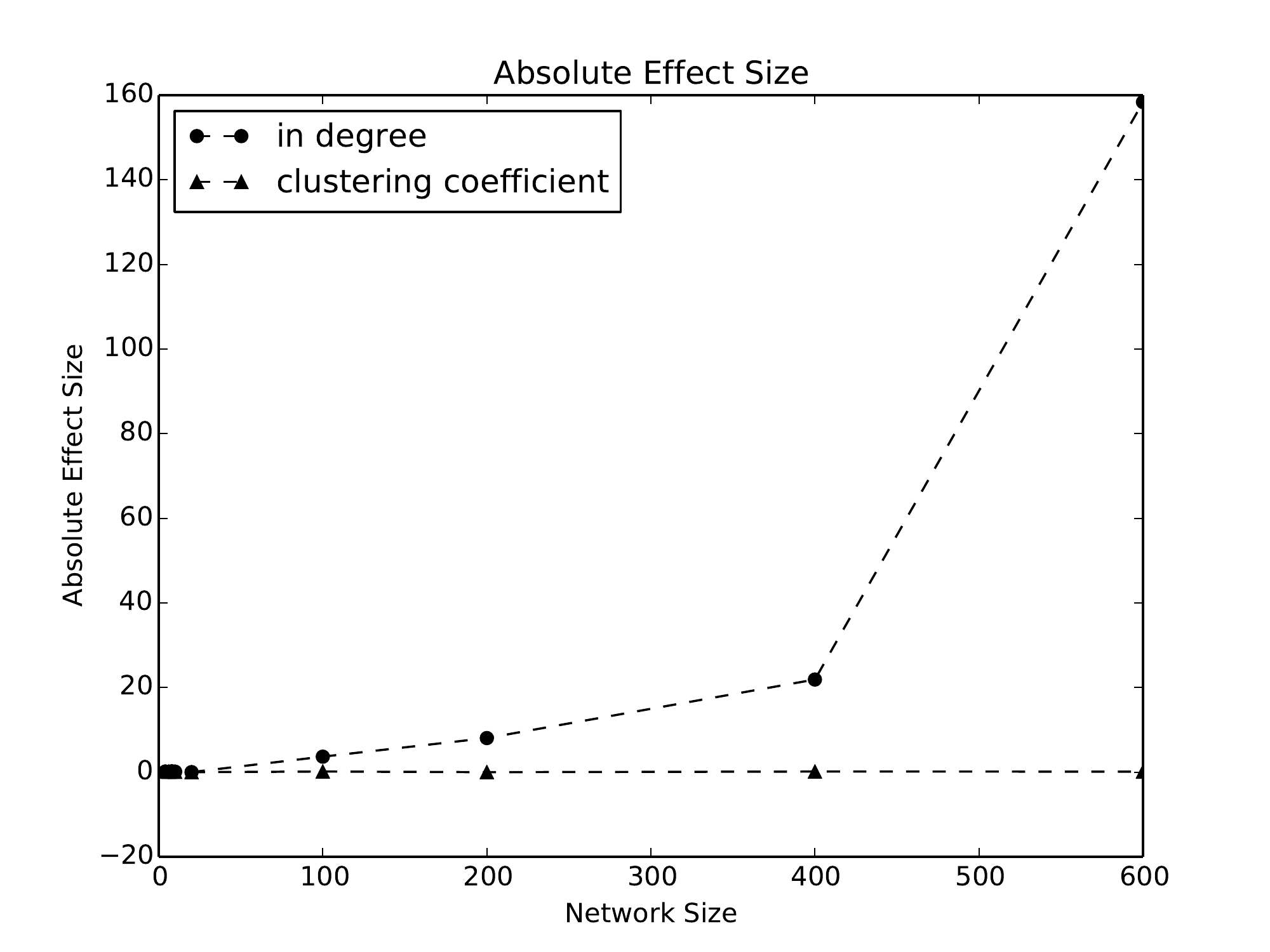}
  \caption{\small The effect size as a function of network size for
    larger networks generated by the deep belief network.}
  \label{fig:effect}
\end{figure*}

\section{Experiments}
\label{sec:experiments}
In order to empirically evaluate the ability of each model to
reproduce sets of networks, we train each model on a set of
ground-truth networks, generate a new set of networks from that model,
and assess whether or not the model reproduced the ground truth using
the techniques described in Section \ref{sec:assessment}.  These
methods all require that we specify a set of network statistics; we
select the in-degree and clustering coefficient statistics, which are
considered to be fundamental properties of networks
\cite{Newman:dbyVK2e2}.  Other network statistics will be
investigated in future work.

\subsection{Synthetic Data}
We compare the relative ability of the candidate ERGMs and the new
models to reproduce network data generated from existing models of
network structure by including a variety of generators with different
underlying dependence assumptions.  The ground-truth models are the
(directed) Erdos-Renyi (ER) model, the configuration model (which
produces graphs with a given degree sequence), and the Krapivsky model
(which produces graphs with dependent marginally heavy-tailed
distributions over in-- and out--degree via preferential attachment)
\cite{Krapivsky:2001vj}.

\subsection{Real Data}
We also assess the ability of each model to reproduce a set of real
networks.  Specifically, we train each model on the Stanford Network
Analysis Project (SNAP) Twitter ego network dataset
\cite{McAuley:2012wy}.  This dataset contains 973 one-hop ego networks
sampled from the online social network Twitter.  Each network is
defined by an `ego' individual, the individuals that they follow, and
the `follows' edges amongst the nodes so defined.  The networks
contain between 6 and 248 nodes and 10 and 18,143 edges, respectively.
This dataset is well-suited to evaluating the models at hand because
it provides networks that are of the same scale which can reasonably
be considered to be independent and identically distributed.

We limit the size of networks considered by sampling from each ego
network. A set of ego networks with $n$ nodes is constructed by, for
all ego nets with at least $n$ nodes, creating a network with the ego
node and $n-1$ randomly chosen nodes and the edges between them.

\section{Results}
\label{sec:results}

\begin{figure}
  \centering
  \textbf{Krapivsky Model}\\
  \includegraphics[scale=0.7]{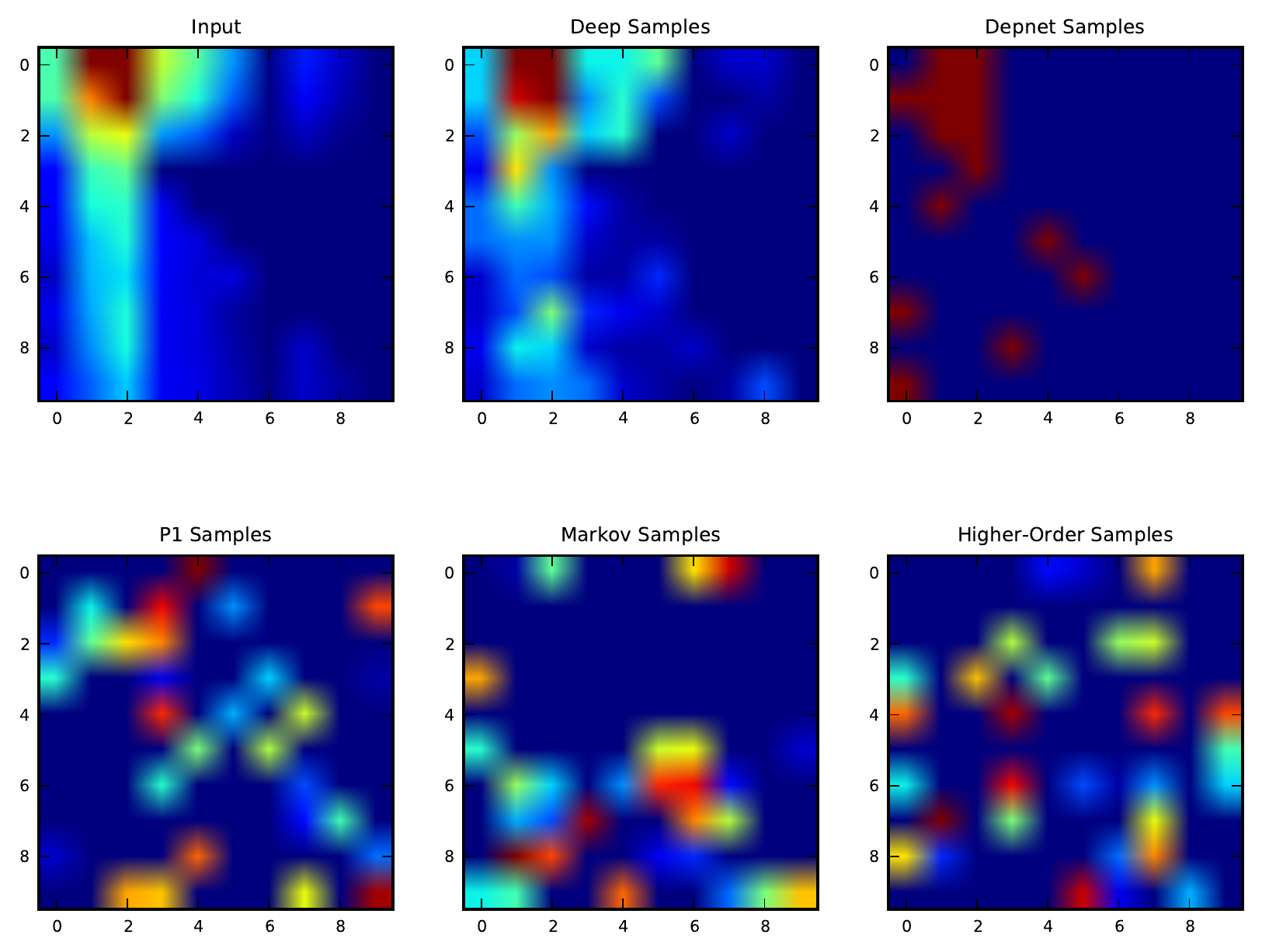}
  \caption{A visual assessment of network reproduction for Krapivsky
    networks with 10 nodes.  Note that the deep model produces an
    average adjacency matrix which most closely resembles the input.}
  \label{fig:krap10}
\end{figure}

We provide results for the directed ER, configuration, and Krapivsky
models Table \ref{tab:synthperm}.  In all cases, we assessed whether
the models reproduced the data with respect to mean in-degree and
clustering coefficient.  200 networks $\netset_1$ were generated from
each ground-truth model and 30 samples $\netset_2$ were generated from
each candidate model.

Table \ref{tab:synthperm} shows the results of applying the
permutation test described in Section \ref{sec:perm} to the synthetic
ground truth $\netset_1$ and model-generated networks $\netset_2$ of
various sizes.  We find that the deep model consistently provides the
best performance; $\netset_1$ and $\netset_2$ are most often found to
not be distinguishable with respect the in-degree or clustering
coefficient.  The the deep model outperforms the p1, Markov, and
higher-order exponential random graph models as well as the dependency
network model.


We were able to apply the deep learner to larger networks than the
other models due to its shorter wall-clock runtime; results are shown in
Table \ref{tab:kraplarge}.  Unfortunately, we found that the deep
model does not maintain its performance for larger networks, and is
almost completely unable to reproduce complex behavior for networks
with more than a handful of nodes.

Note that the decreasing p-values seen in Table \ref{tab:kraplarge}
could be natural consequence of increasing statistical power rather
than an indication that the deep model has failed to learn complex
behavior.  To assess this, we plotted the absolute effect size as a
function of network size, as shown in Figure \ref{fig:effect}.  The
sharp increase in the effect size for the in-degree distribution
suggests that the deep model is not learning to generate larger
networks with correct degree distributions.  On the other hand, he
flat effect size for clustering coefficient suggests that the deep
model is learning to generate larger networks with reasonable
distributions of clustering coefficients.  This may explain why the
deep model is able to pass the clustering coefficient permutation test
for networks with 200 nodes.

Figure \ref{fig:krap10} shows a visual assessment of each model's
reproduction of networks with 10 nodes generated from the Krapivsky
models.  The heatmaps display the average adjacency matrix of the
ground-truth set and the adjacency matrixes sampled from each model.
Strikingly, the deep model's reproduction is very similar in character
to the input; conversely, the average adjacency matrices produced by
the exponential random graph models look almost nothing like the
input.  This is at least partially due to an important difference
between the model classes: the deep model encodes \emph{particular
  edges}, whereas the ERGMs we consider, working only with a
distribution over configurations, do not.  The deep model will
accordingly provide much better results when the task entails learning
the structure of labeled edges.

\section{Conclusion}
\label{sec:conclusion}

We consider whether deep belief networks, dependency networks, and
members of the exponential random graph family can learn to generate
networks whose complex behavior is consistent with a set of input
examples. We find that deep belief networks are the only model to
perform well at any network size, but that all models perform poorly
when generating moderate to large networks.

\bibliography{nips2014} \bibliographystyle{unsrt}

\end{document}